%% file: icml_main.tex
\documentclass{article}

\usepackage{microtype}
\usepackage{graphicx}
\usepackage{subcaption}
\usepackage{booktabs} % for professional tables

\usepackage{hyperref}

\usepackage[preprint]{icml2026}

\usepackage{amsmath}
\usepackage{amssymb}
\usepackage{mathtools}
\usepackage{amsthm}

\usepackage[capitalize,noabbrev]{cleveref}

\theoremstyle{plain}
\newtheorem{theorem}{Theorem}[section]
\newtheorem{proposition}[theorem]{Proposition}

\theoremstyle{definition}
\newtheorem{definition}[theorem]{Definition}

\theoremstyle{remark}

\usepackage[textsize=tiny]{todonotes}

\icmltitlerunning{Learning Tractable Distributions of Language Model Continuations}

\input{math_commands.tex}

\usepackage{tikz}
\usetikzlibrary{positioning,arrows.meta,decorations.pathreplacing,shapes.misc}
\usepackage{multirow}
\usepackage{hyperref}
\usepackage{url}
\usepackage{booktabs}
\usepackage{graphicx}
\usepackage{subcaption}
\usepackage[rgb, dvipsnames,table]{xcolor}

\newtheorem{example}{Example}

\usepackage{pifont}
\newcommand{\cmark}{\textcolor{green}{\normalsize \ding{51}}}%
\newcommand{\xmark}{\textcolor{red}{\normalsize \ding{55}}}%

\usepackage{cleveref}
\usepackage{tikz}
\usetikzlibrary{calc}
\usepackage{wrapfig}
\usepackage{bbm}

\definecolor{lightskyblue}{rgb}{0.53, 0.81, 0.98}

\begin{document}

\twocolumn[
  \icmltitle{Learning Tractable Distributions of Language Model Continuations}



  \icmlsetsymbol{equal}{*}

% \author{Gwen Yidou-Weng$^1$, Ian Li$^2$, Anji Liu$^3$, Oliver Broadrick$^1$, Guy Van den Broeck$^1$, Benjie Wang$^1$ \\
% $^1$University of California, Los Angeles\\
% $^2$University of California, San Diego\\
% $^3$National University of Singapore
% }
  \begin{icmlauthorlist}
    \icmlauthor{Gwen Yidou-Weng}{1}
    \icmlauthor{Ian Li}{2}
    \icmlauthor{Anji Liu}{3}
    \icmlauthor{Oliver Broadrick}{1}
    \icmlauthor{Yuchen Cui}{1}
    \icmlauthor{Guy Van den Broeck}{1}
    \icmlauthor{Benjie Wang}{1}
  \end{icmlauthorlist}

  \icmlaffiliation{1}{University of California, Los Angeles}
  \icmlaffiliation{2}{University of California, San Diego}
  \icmlaffiliation{3}{National University of Singapore}

  \icmlcorrespondingauthor{Gwen Yidou-Weng}{gwenweng@ucla.edu}

  \icmlkeywords{ keywords todo }

  \vskip 0.3in
]



\printAffiliationsAndNotice{}  % no special notice (required even if empty)

% The \author macro works with any number of authors. There are two commands
% used to separate the names and addresses of multiple authors: \And and \AND.
%
% Using \And between authors leaves it to \LaTeX{} to determine where to break
% the lines. Using \AND forces a linebreak at that point. So, if \LaTeX{}
% puts 3 of 4 authors names on the first line, and the last on the second
% line, try using \AND instead of \And before the third author name.

% \newcommand{\fix}{\marginpar{FIX}}
% \newcommand{\new}{\marginpar{NEW}}

\input{macros}

\begin{abstract}
Controlled generation imposes sequence-level constraints (syntax, style, safety) that depend on future tokens, making exact conditioning of an autoregressive LM intractable. Tractable surrogates such as HMMs can approximate continuation distributions and steer decoding, but standard surrogates are often weakly context-aware. We propose \emph{Learning to Look Ahead} (\textsc{LTLA}), a hybrid method that uses base-LM embeddings to condition a \emph{globally learned} tractable surrogate: a neural head predicts only a prefix-dependent latent prior, while a shared HMM answers continuation queries exactly. LTLA is designed to avoid two common efficiency traps when adding neural context. First, it avoids vocabulary-sized prefix rescoring ($V$ extra LM evaluations) by scoring all next-token candidates via a single batched HMM forward update. Second, it avoids predicting a new HMM per prefix by learning one shared HMM and conditioning only the latent prior, which enables reuse of cached future-likelihood (backward) messages across decoding steps. Empirically, \textsc{LTLA} improves continuation likelihood over standard HMM surrogates, enables lookahead control for vision--language models by incorporating continuous context, achieves 100\% syntactic constraint satisfaction, and improves detoxification while adding only a 14\% decoding-time overhead.
\end{abstract}

\section{Introduction}
% \gw{thanks everyone for input! the major change was to emphasize whats interesting about our method: instead of simply neural + hmm and boom! it works, we point out the challenges in adding neural to TPM and our design choice accordingly.

Autoregressive models are the dominant way to represent high-dimensional discrete distributions over language and sequence data \citep{grattafiori2024llama,shin2021protein}, factorizing the distribution into a sequence of next-token conditional distributions. Many useful queries, however, concern properties of the entire sequence: to generate autoregressively under a sequence-level constraint, we need the probability that the remaining suffix will satisfy that constraint given the current prefix \citep{boyd2022predictive,ZhangICML23}. Computing this probability requires reasoning over exponentially many possible futures and is therefore intractable for standard autoregressive language models (LM). In practice, estimates come from sampling \citep{qin2022cold,hu2023amortizing,lew2023sequential}, which introduces significant computational overhead, or from learned heuristics \citep{krause2020gedi,yang2021fudge,meng2022controllable} that must be tailored to each query or constraint.

\begin{figure*}
    \centering
    \begin{subfigure}[t]{0.47\linewidth}
        \centering
        \includegraphics[width=\linewidth]{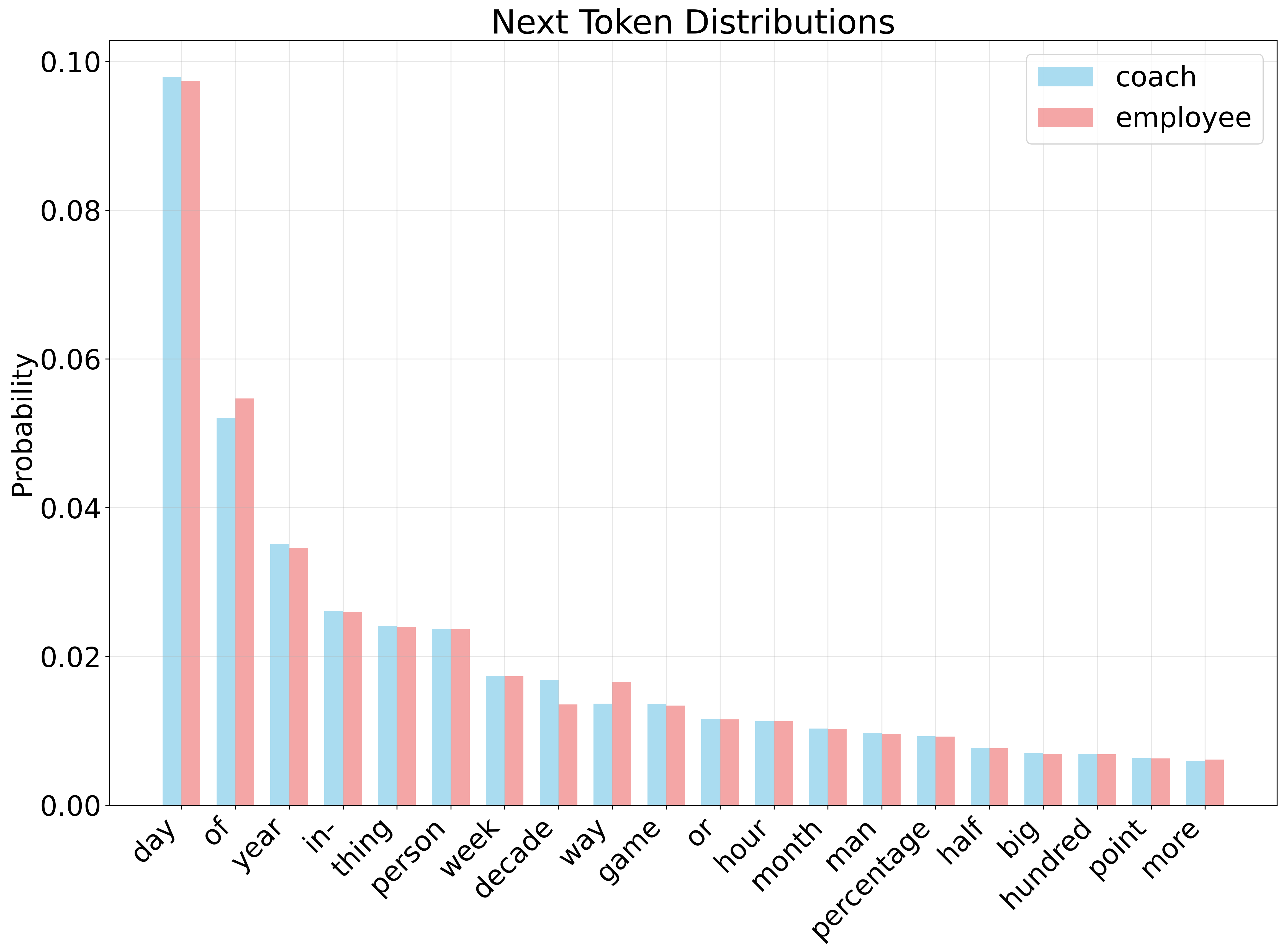}
        \vspace{-16pt}
        \caption{Standard HMM}
    \end{subfigure}
    \begin{subfigure}[t]{0.47\linewidth}
        \includegraphics[width=\linewidth]{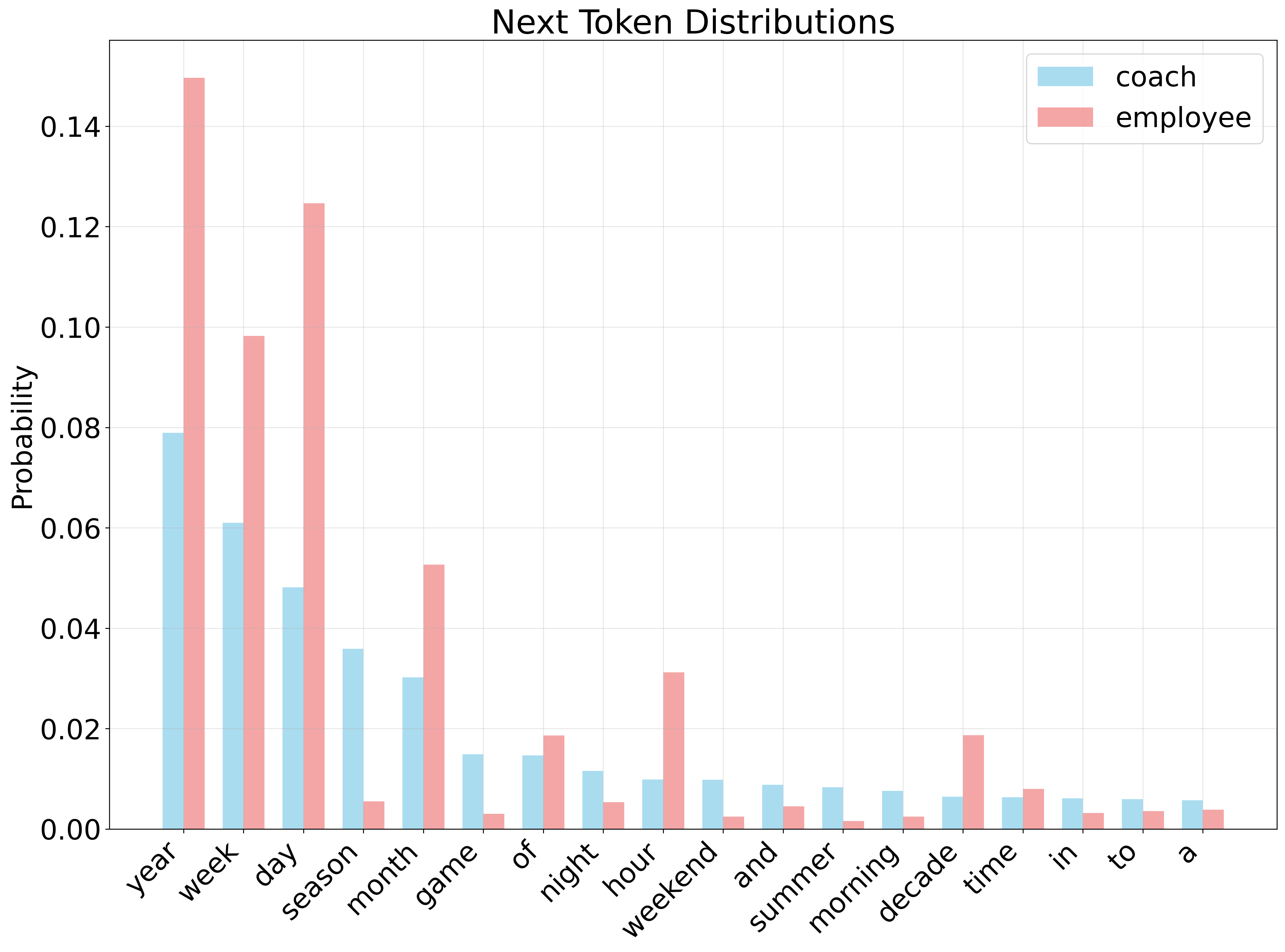}
        \vspace{-16pt}
        \caption{Neural HMM (LTLA)}
    \end{subfigure}
    \vspace{-4pt}
    \caption{The encoder of standard HMMs is often insensitive to information contained within the context. For example, with the context \texttt{they fired the <x> after just one}, where \texttt{<x>} can be \texttt{coach} or \texttt{employee}, the distributions are almost identical for the standard HMM, while our neural HMM shows a significant shift in distribution (e.g, with \texttt{season} and \texttt{game} more likely for \texttt{coach}). 
    % \anji{Maybe we can make this figure more vivid (e.g., with some thought bubbles with a few plausible \textbf{multi-token continuations}).}
    }
    \label{fig:motivation}
\end{figure*}

Tractable probabilistic models (TPMs) are a broad class of generative models that admit efficient routines for computing such conditional queries exactly \citep{ProbCirc20}. As such, a natural approach employed by prior works \citep{deng2025decoupling,ZhangICML23,ZhangNeurIPS24,YidouWengICML25} is to utilize TPMs as \emph{tractable surrogates} to the LM, such that queries on the LM can be approximated by computing them on the surrogate. 
%In particular, conditional queries are executed by utilizing a probabilistic \emph{encoder} to summarize the prefix/context, and a probabilistic \emph{decoder} that can tractably compute the conditional probability given the encoded prefix/context.
However, TPMs commonly used for language modeling, such as hidden Markov models (HMMs), face two obstacles in this role. First, we find empirically that standard HMMs are often insensitive to prefix information, leading to less accurate continuation distributions. For example, in Fig.~\ref{fig:motivation}, changing \texttt{<x>} from \texttt{coach} to \texttt{employee} barely changes the HMM's next-token distribution, despite the clear semantic change in the prefix. Second, as discrete sequence models, they do not naturally incorporate continuous context, such as image embeddings produced by modern vision--language models (VLMs).

Our key insight is that, to encode past context effectively and answer future queries efficiently, these two tasks should be decoupled. Understanding the past (\emph{lookback}) should preserve as much information as possible, whereas forecasting the future (\emph{lookahead}) must remain tractable. To achieve this, we reuse the base LM's rich representation of the prefix to condition a tractable probabilistic inference model (e.g., an HMM) over continuations, yielding a lookahead distribution that is sensitive to semantic differences in the context (e.g., shifting mass toward \texttt{season} and \texttt{game} when \texttt{<x>}=\texttt{coach} in Fig.~\ref{fig:motivation}). In this work, we propose a \emph{hybrid} tractable model that uses a transformer-based LM for lookback and an HMM for lookahead.

Making this hybrid model practical raises two concrete challenges. First, when autoregressively generating the next token, a naïve strategy would run the transformer on the prefix concatenated with every candidate token across the vocabulary to assess how each affects the downstream conditional distribution; this per-step sweep over the entire vocabulary is expensive \emph{(see “Exhaustive LM Rescoring” in Table~\ref{tab:approach-comparison})} and conceptually corresponds to a branching tree of continuations, as shown in Fig.~\ref{fig:ltla-lookahead}(a). Second, conditioning the surrogate itself on the prefix by predicting fresh parameters for every context remains tractable at a single step but prevents reuse: as the prefix grows, the surrogate must be rebuilt and its future probabilities recomputed at every step, which increases decode-time cost and memory \emph{(see “Prefix-Parameterized TPM” in Table~\ref{tab:approach-comparison}; cf. Sec.~\ref{sec:only latent conditional})}. Both issues undercut the efficiency that makes TPMs attractive.

Thus, we encode rich context with the transformer while preserving TPM efficiency by conditioning only the HMM surrogate's latent-state prior on the prefix, and using a single globally learned HMM shared across contexts. Specifically, the transformer's hidden states are fed into a lightweight head that is trained to produce a prior over the surrogate’s latent state, and the observed next token $x_t$ is incorporated with a one-step HMM update as a single batched matrix–vector operation, avoiding separate computation for each token in the vocabulary. With fixed transition and emission parameters, future computations are reusable across steps, so conditional queries remain exact and fast even for long sequence lengths \emph{(see “LTLA” in Table~\ref{tab:approach-comparison})}. Fig.~\ref{fig:ltla-lookahead}(b) illustrates this design: the LM handles lookback, while a linear-chain HMM surrogate performs lookahead with a single dynamic program. We call this approach \emph{Learning to Look Ahead (LTLA)}. %LTLA thus provides a tractable lookahead model that improves upon existing TPMs while incurring minimal overhead over the base language model. 
 %reuses LM activations with minimal overhead, and applies in multimodal settings where a standalone TPM cannot encode continuous context.

%Empirically, relative to an unconditional HMM that struggles to separate prefixes, LTLA achieves higher conditional log-likelihood, especially at short horizons. It improves constraint satisfaction at comparable fluency on controlled-generation tasks for both language and vision–language models, while adding only a small inference-time overhead.

Our contributions can be summarized as follows:

1. We introduce \textsc{LTLA}, a hybrid approach for learning tractable surrogates to language models. \textsc{LTLA} reuses the base LM to predict a prefix-conditioned latent prior for an HMM over continuations, enabling accurate conditional queries at each generation step with only a small number of matrix--vector operations. This prefix conditioning makes the surrogate more context-sensitive, producing sharper shifts toward compatible continuations (e.g., Fig.~\ref{fig:motivation}), and it extends naturally to multimodal context in VLMs.

2. We study architectural choices for both the neural encoder and the tractable decoder, and quantify their trade-offs in surrogate accuracy and decoding overhead. In our experiments, a lightweight linear head already captures most of the gains while adding negligible overhead, so we use it as the default configuration.

3. Empirically, \textsc{LTLA} improves conditional log-likelihood over standard tractable surrogates, reducing perplexity by roughly 10--15\% across hidden sizes. The largest gains occur in the next few tokens: next-token perplexity drops by 60\%. On controlled generation, \textsc{LTLA} achieves 100\% syntactic constraint satisfaction (e.g., CommonGen) and substantially improves semantic control in detoxification, reducing average toxicity from 0.254 to 0.015, while adding only a 14\% inference-time overhead.

\begin{figure*}[t]
\centering
\scriptsize
\begin{tikzpicture}[
    font=\scriptsize,
    >=Latex,
    node distance=5mm and 8mm,
    box/.style={
        draw, rounded corners, thick, align=center,
        minimum width=.6cm,
        minimum height=0.65cm
    },
    smallbox/.style={
        draw, rounded corners, thick, align=center,
        minimum width=.7cm,
        minimum height=0.6cm
    },
    token/.style={
        circle, draw, thick,
        minimum size=7mm,
        inner sep=0pt
    },
    latent/.style={
        circle, draw, thick,
        minimum size=7mm,
        inner sep=0pt,
        fill=gray!5
    },
    obs/.style={
        circle, draw, thick,
        minimum size=7mm,
        inner sep=0pt
    },
    brace/.style={
        decorate,
        decoration={brace, amplitude=3pt}
    },
    every path/.style={thick}
]

% =========================
% Panel (a): Naive LM lookahead (left)
% =========================

\begin{scope}[xshift=0cm]

\node[anchor=west] at (-3.8,2.4) {\textbf{(a) Naive lookahead}};

\node[box, fill=gray!10] (lmroot) at (-3.8,0.9) {LM};

\node[token, right=4mm of lmroot] (xt) {$x_t$};
\draw[->] (lmroot) -- (xt);

\node[token, above right=4mm and 12mm of xt]  (cand1) {$x_{t+1}^{(1)}$};
\node[token, right=24mm of xt]               (cand2) {$x_{t+1}^{(2)}$};
\node[token, below right=4mm and 12mm of xt] (cand3) {$x_{t+1}^{(3)}$};

\draw[->] (xt) -- (cand1);
\draw[->] (xt) -- (cand2);
\draw[->] (xt) -- (cand3);

\node[token, above right=1mm and 10mm of cand1] (cand11) {$x_{t+2}^{(1)}$};
\node[token, below right=-3mm and 18mm of cand1] (cand12) {$x_{t+2}^{(2)}$};

\node[token, above right=0mm and 18mm of cand2] (cand21) {$x_{t+2}^{(3)}$};
\node[token, below right=0mm and 18mm of cand2] (cand22) {$x_{t+2}^{(4)}$};

\node[token, above right=-3mm and 18mm of cand3] (cand31) {$x_{t+2}^{(5)}$};
\node[token, below right=1mm and 10mm of cand3]  (cand32) {$x_{t+2}^{(6)}$};

\foreach \p/\c in {cand1/cand11,cand1/cand12,cand2/cand21,cand2/cand22,cand3/cand31,cand3/cand32} {
    \draw[->] (\p) -- (\c);
}

% \draw[]
%     ([xshift=-120mm,yshift=3mm]cand11.east) --
%     ([xshift=30mm,yshift=-3mm]cand32.east)
%     node[midway,xshift=20mm,align=left]
%     {\footnotesize LM forward pass\\[-1pt]\footnotesize for each path};

% \node at (-3.8,1.5) {\footnotesize LM forward pass\\[-1pt]\footnotesize for each path};

\node[anchor=west,align=left] at (-3.8,-.5) {\footnotesize LM forward pass\\ \footnotesize for each path};

\end{scope}

% =========================
% Panel (b): LTLA lookahead (right)
% =========================

\begin{scope}[xshift=8cm]

\node[anchor=west] at (-3.8,2.4) {\textbf{(b) LTLA lookahead}};

\node[box, fill=gray!10] (lm2root) at (-3.8,1.2) {LM};

\node[
    rectangle, draw, rounded corners,
    minimum width=.6cm,
    minimum height=.7cm,
    right=of lm2root,
    fill=blue!5
] (ht) {$h_t$};
\draw[->] (lm2root) -- (ht);

\node[smallbox, fill=blue!10, right=of ht] (head) {head};
\draw[->] (ht) -- (head);

\node[latent, right=5mm of head] (zt) {$z_t$};
\node[latent, right=5mm of zt]   (zt1) {$z_{t+1}$};
\node[latent, right=5mm of zt1]  (zt2) {$z_{...}$};
\node[latent, right=5mm of zt2]  (ztn) {$z_{t+n}$};

\draw[->] (head) -- (zt);
\draw[->] (zt) -- (zt1);
\draw[->] (zt1) -- (zt2);
\draw[->] (zt2) -- (ztn);

\node[obs, below=6mm of zt]  (xt0) {$x_t$};
\node[obs, below=6mm of zt1] (x1)  {$x_{t+1}$};
\node[obs, below=6mm of zt2] (x2)  {$x_{...}$};
\node[obs, below=6mm of ztn] (xn)  {$x_{t+n}$};

\draw[->,orange!80!black] (zt)  -- (xt0);
\draw[->,orange!80!black] (zt1) -- (x1);
\draw[->,orange!80!black] (zt2) -- (x2);
\draw[->,orange!80!black] (ztn) -- (xn);

\node[anchor=west,align=left] at (-3.8,-.5) {\footnotesize One LM forward pass;\\ \footnotesize then tractable HMM inference};

\end{scope}

\end{tikzpicture}
\vspace{-.5em}
\caption{{\textbf{Naive LM lookahead vs LTLA.}
  \textbf{(a)} Naive lookahead branches over many candidate continuations and requires a separate LM forward pass for each path, which is exponential as the horizon grows.
  \textbf{(b)} LTLA runs the LM once on the prefix to obtain a context embedding $h_t$, maps $h_t$ through a lightweight head to a prior over latent states $z_t$, and then uses an HMM to model future continuations. Lookahead scores are obtained by a single linear dynamic program instead of repeated LM calls.}}
\label{fig:ltla-lookahead}
\end{figure*}

\begin{table*}[t]
\small
\centering
\caption{Comparison of lookahead properties across models.}

\label{tab:approach-comparison}
\begin{tabular}{lccccc}
\toprule
\textbf{Model} &
\begin{tabular}[c]{@{}c@{}}\textbf{Tractable}\\ \textbf{Lookahead}\end{tabular} &
\begin{tabular}[c]{@{}c@{}}\textbf{Context Awareness}\\ \textbf{(incl.\ multimodal)}\end{tabular} &
\begin{tabular}[c]{@{}c@{}}\textbf{No Extra LM}\\ \textbf{Calls per Step?}\end{tabular} &
\begin{tabular}[c]{@{}c@{}}\textbf{Reuse Surrogate}\\ \textbf{Precompute per Prefix?}\end{tabular} &
\begin{tabular}[c]{@{}c@{}}\textbf{Decoding}\\ \textbf{Overhead}\end{tabular} \\
\midrule
LLM & \xmark & \cmark & --- & --- & --- \\
Standard HMM & \cmark & \xmark & \cmark & \cmark & \textbf{Low} \\
Exhaustive LM Rescoring & \xmark & \cmark & \xmark & \cmark & High \\
Prefix-Parameterized TPM & \cmark & \cmark & \cmark & \xmark & High \\
\textbf{LTLA} & \cmark & \cmark & \cmark & \cmark & \textbf{Low} \\
\bottomrule
\end{tabular}
\end{table*}

% \guy{We should add some language to address the question: why is this difficult? Why not make the HMM parameters a function of the context? What was challenging in designing this solution?}

\section{Tractable Modeling of Sequences}

We study autoregressive sequence models, and how to effectively answer \emph{queries} about the distribution they represent. An autoregressive model decomposes the distribution over a sequence of tokens $x_{1:T}$ as 
% \begin{equation}
\(p(x_{1:T}) = \prod_{t=1}^T p(x_{t} | x_{<t}).\)
% \end{equation}
\paragraph{Queries} Aside from generating or analyzing the next token distribution, we are often interested in more complex properties of the distribution. These can be represented generally as \emph{conditional probability queries} \citep{boyd2022predictive}, which ask for the probability $p(\alpha | x_{1:t})$ of some event $\alpha$, where $x_{1:t}$ is the prefix (or \emph{context}) generated so far. Examples of events $\alpha$ might include the $k^{\text{th}}$ token $x_{t+k}$ in the future taking a particular value, some token $a$ appearing before token $b$ in the sequence, the expected length of the sequence generated, or more complex properties involving grammatical or semantic constraints \citep{ZhangNeurIPS24,YidouWengICML25,ahmed2025semantic, deng2025decoupling}. What these queries have in common is that they require \emph{looking into the future}: that is, aggregating over all possible continuations, weighted by their conditional probability given the context:
\begin{equation} \label{eq:cond_query}
    p(\alpha | x_{1:t}) = \sum_{x_{t+1:T}} p(x_{t+1:T} | x_{1:t}) p(\alpha | x_{1:t}, x_{t+1:T}).
\end{equation}
One of the key downstream applications of conditional probability queries is \emph{controlled generation}: that is, generating from an autoregressive model \emph{conditional on} some event $\alpha$. In particular, 
observe that, by Bayes' rule, the distribution of the next token $X_t$ conditional on previous tokens and $\alpha$ is given by:
\begin{equation} \label{eq:control}
p(X_t|x_{<t},\alpha)\propto p(X_t|x_{<t})\cdot p(\alpha|x_{< t}, X_t)
\end{equation}
As such, if one has access to an oracle for conditional probability queries, then it is possible to sample autoregressively from the conditional distribution by explicitly computing the terms in Equation~\ref{eq:control}.
In practice, however, such an oracle is not available, and one must resort to approximations.

\paragraph{Estimating Conditional Probability Queries} Consider the problem of estimating the conditional probability query $p(\alpha | x_{1:t}) $ given in Equation \ref{eq:cond_query}. 
The key tradeoff is between (i) the \emph{accuracy} and (ii) the \emph{computation cost} of the estimation.
Explicitly enumerating all such continuations would result in an exact answer, but is clearly infeasible as the number of such continuations grows exponentially with sequence length. As such, for most models, one typically needs to approximate, for example by (i) using sampling-based techniques targeting the conditional continuation distribution \citep{qin2022cold,zhao2024probabilistic,loula2025syntactic}; or (ii) directly approximating the conditional probability query using a neural classifier or generative model specialized to the constraint $\alpha$ \citep{krause2021gedi,yang2021fudge,meng2022controllable}.

%As such, one must resort to sampling from the autoregressive model's distribution to obtain a Monte Carlo estimate. 

In this work, we consider an alternative, computationally efficient approach based upon \emph{tractable modeling} of continuations. This stems from the observation that, for certain distributions $p$ and queries $\alpha$, the computation of $p(\alpha|x_{1:t})$ can be done both (i) exactly and (ii) efficiently, breaking the tradeoff. In particular, tractable probabilistic models (TPMs) \citep{ProbCirc20} are classes of probabilistic generative models that are known to enable computationally efficient \textit{analytical} computation of many classes of queries, such as marginal probabilities.

\begin{example}
 Hidden Markov models (HMMs) are tractable sequence models, that represent a joint distribution over a sequence of $T$ variables $X_1,X_2,\ldots,X_T$ each taking values in a size-$V$ vocabulary $\mathcal{V} = \{0, \ldots, V-1\}$ with latent variables $Z_1,Z_2,\ldots,Z_T$ each taking value in a discrete set of hidden states $\mathcal{H} = \{0, \ldots, H-1\}$ of size $H$. The parameters of an HMM are given by an emission matrix $q(x_t| z_t) \in \mathbb{R}^{H \times V}_{\geq 0}$ and a transition matrix $q(z_{t} | z_{t-1}) \in \mathbb{R}^{H \times H}_{\geq 0}$. Then, the distribution of an HMM is defined by 
\begin{align*}
&q(x_{1:T})
% \\&
=\sum_{z_1,\ldots,z_T}q(z_1)q(x_1|z_1)\prod_{t=2}^T q(z_t|z_{t-1})q(x_t|z_{t}).
\end{align*}

HMMs enable efficient computation of various queries via (variants of) the \emph{forward} and \emph{backward} algorithms \citep{rabiner1986introduction}. For example, if $\alpha$ is the event that the last token $X_T = \texttt{world}$, we can compute the conditional query $q(\alpha|x_{1:t})$ using the fact that (by conditional independence):
\begin{equation} \label{eq:forward_backward}
    q(\alpha | x_{1:t}) = \sum_{z_t} q(z_t | x_{1:t}) q(\alpha|z_t).
\end{equation}
where $q(z_t | x_{1:t})$ and $q(\alpha | z_t)$ can be computed using the forward and backward algorithms respectively, which each take linear time in the sequence length. For instance, $q(\alpha | z_t)$ can be computed using the following recurrence relation backward in time:
\begin{equation}
    q(\alpha|z_{t-1}) = \sum_{z_{t}} q(z_t|z_{t-1}) q(\alpha |z_t)
\end{equation}
with base case $q(\alpha|z_T) = q(x_T=\texttt{world}|z_T)$. Prior work has also shown that HMMs support tractable querying of many other conditions $\alpha$, including complex logical constraints\footnote{Specifically, constraints represented as deterministic finite automata (DFA) or unambiguous context-free grammars (uCFG).} \citep{ZhangNeurIPS24,zhang2025restructuring,marzouk2022marginal, deng2025decoupling} or factorized classifiers for semantic constraints \citep{vergari2021compositional,YidouWengICML25}. 
\end{example}

% and $q(\alpha) = \sum_{z_1}q(z_1) q(\alpha|z_1)$.

The LLM distribution $p$ we are interested in will not be tractable in this way. However, we can aim to approximate the (prior) distribution $p(x_{t+1:T} | x_{1:t})$ using a simpler tractable surrogate model $q(x_{t+1:T} | x_{1:t})$ for which conditional queries are tractable, and estimating $p(\alpha | x_{1:t}) \approx q(\alpha|x_{1:t})$ using the tractable approximation $q$. 
%In more detail, we consider model classes $\mathcal{Q}$ specifying distributions $q(X_{t+1:n})$ over continuations $X_{t+1:n}$ such that conditional probability queries of the form $q(\alpha) = \sum_{x_{t+1:n}} q(x_{t+1:n}) q(\alpha | x_{t+1:n}, x_{1:n})$ are tractable.
%Then, if we can approximate the true autoregressive model's conditional distribution $p(x_{t+1:n}|x_{1:n})$ with some $q \in \mathcal{Q}$, then we can estimate $p(\alpha | x_{1:n}) \approx q(\alpha)$. 
This approach has been employed in prior works \citep{deng2025decoupling,ZhangNeurIPS24,YidouWengICML25} using HMMs as the TPM of choice, and has been shown to lead to state-of-the-art performance on controlled generation benchmarks.

%In this work, we consider an alternative, computationally efficient approach based upon \emph{tractable modeling} of continuations. 
%Tractable probabilistic models (TPMs) \citep{ProbCirc20} are classes of probabilistic generative models that enable computationally efficient \textit{analytical} computation of certain classes of queries, such as marginal probabilities. Intuitively speaking, instead of attempting to directly approximate the \emph{posterior} distribution $p(x_{t+1:n} | x_{1:t}, \alpha)$, as other methods do, we aim to approximate the \emph{prior} distribution $p(x_{t+1:n} | x_{1:t})$ using a simpler model $q(x_{t+1:n})$ for which conditioning on $\alpha$ can be done tractably.
%In more detail, we consider model classes $\mathcal{Q}$ specifying distributions $q(X_{t+1:n})$ over continuations $X_{t+1:n}$ such that conditional probability queries of the form $q(\alpha) = \sum_{x_{t+1:n}} q(x_{t+1:n}) q(\alpha | x_{t+1:n}, x_{1:n})$ are tractable.
%Then, if we can approximate the true autoregressive model's conditional distribution $p(x_{t+1:n}|x_{1:n})$ with some $q \in \mathcal{Q}$, then we can estimate $p(\alpha | x_{1:n}) \approx q(\alpha)$. This approach has been employed in prior works \citep{ZhangICML23} using Hidden Markov models (HMMs) as the TPM of choice.

%\paragraph{The Promise and Challenges of Tractable Modeling}  

%\paragraph{The Promise and Challenges of Tractable Modeling}  

%\paragraph{Desiderata for Tractable Modeling} 

%Let us also consider the computational cost during 

% \vspace{-16pt}
\paragraph{Promise and Challenges of Tractable Surrogates} 
%\paragraph{Tractable Surrogate Models}

Besides state-of-the-art empirical performance on downstream applications, the tractable modeling approach offers a number of other benefits related to computational efficiency. Firstly, it amortizes the cost across different constraints~$\alpha$. That is, since $q$ is trained to match the \emph{prior}, and the computation of the query $q(\alpha|x_{1:t})$ is conducted using a symbolic algorithm, one does not need to commit in advance to any particular condition $\alpha$. This is in contrast to approaches that train models to specifically target the \emph{posterior} $p(x_{t+1:n} | x_{1:t}, \alpha)$.
%Prior work has shown that a general class of TPMs known as probabilistic circuits support tractable querying of many conditions $\alpha$, including conditioning on complex logical constraints\footnote{more specifically, deterministic finite automata (DFA) or unambiguous context-free grammars (uCFG).} \citep{ZhangNeurIPS24,zhang2025restructuring}  or factorized classifiers for semantic constraints \citep{YidouWengICML25}. 
Secondly, tractable models can amortize the cost across different \emph{contexts} $x_{1:t}$ by exploiting conditional independence. For example, in the HMM computation in Equation \ref{eq:forward_backward}, the backward quantity $q(\alpha|z_t)$ can be precomputed and cached independently of the context~$x_{1:t}$.

%the computational cost of using tractable models is typically much lower than other approaches that require multiple evaluations of large models at decoding time. In particular, 

%the use of a tractable model enables computing $q(\alpha|x_{< t}, X_t)$ 
%for all values of $X_t$ simultaneously without additional computation, which drastically reduces the computational cost of using it for controllable generation. %That is, one can train $q$ without any knowledge of $\alpha$, and only introduce $\alpha$ during computation of the query. Secondly, 

%The key advantage of using tractable models of continuations such as HMMs is in its computational efficiency. Consider, for instance, the task of computing $q(\alpha|x_{< t}, X_t)$ for all values of $X_t$, as used in controllable generation. One can precompute the cache $q(\alpha | z_t)$ \emph{independently of the context} $x_{1:t}$. To obtain 
%As such,  compared to a naive conditional query oracle where we would need $O(VC)$ calls for $C$ contexts (one for each context and token $X_t$), the HMM requires just a single computation
%\bw{add more here perhaps}

%owever, a key challenge is to 

This computational efficiency, however, also comes at a cost. The quality of the query estimate provided by the tractable model depends significantly on the quality of the approximation of $q$ to $p$, which in turn has been shown to affect downstream performance \citep{ZhangICML23,YidouWengICML25, deng2025decoupling}. %As we showed in Figure \ref{fig:motivation}, The quality of the estimate obtained 
Unfortunately, the class of tractable models (e.g., HMMs) is fundamentally less expressive than e.g., neural autoregressive models \citep{choi2019relative,BroadrickICML25}.
This is reflected in our observation in Figure~\ref{fig:motivation}, where a HMM trained to approximate a GPT2-large language model is unable to effectively encode dependence on context $x_{1:t}$ in its distribution $q(x_{t+1:n}|x_{1:t})$. As such, a key challenge is to improve the expressivity and learning performance of the tractable model approximation, while maintaining the computational efficiency of the existing HMM-based approximation. In the next section, we will present our approach, Learning To Look Ahead (LTLA), which utilizes an amortized inference approach with a neural encoder to obtain the tractable approximation $q$.

\section{Learning to Look Ahead}

\newcommand{\LM}{\text{LM}}

To formalize our problem from first principles, suppose that we have a set of contexts $\{x_{1:t}^{(j)} \}_{j=1}^{N}$, and let $\mathcal{Q}$ be a class of tractable distributions over continuations (e.g. HMMs). Then our goal is to infer, for each given context $x_{1:t}^{(j)}$, a distribution $q^{(j)} \in \mathcal{Q}$ over continuations $x_{t+1:n}$ such that $q^{(j)} \approx p(\cdot | x_{1:t}^{(j)})$, for example, by maximizing conditional log-likelihood 
\begin{equation}
     \text{LL} = \mathbb{E}_{x_{t+1:T} \sim p(\cdot | x_{1:t}^{(j)})}\left[\log q^{(j)}(x_{t+1:T})\right].
\end{equation}
Unfortunately, in practice we cannot afford to optimize $q^{(j)}$ separately for every context $x_{1:t}^{(j)}$; instead we must take an amortized inference approach, in which we \emph{learn} to predict $q(x_{t+1:T})$ given context $x_{1:t}$. In particular, an HMM trained on the joint distribution over contexts and continuations can be viewed as performing amortized inference by (i) applying a probabilistic \emph{encoder} $q(z_t|x_{1:t})$ to predict the latent state distribution and (ii) parameterizing the distribution over continuations via this latent state and the probabilistic \emph{decoder} $q(x_{t+1:T}|z_t)$. Our key insight is that, for answering queries about the continuation, we only need to be able to (i) evaluate the encoder and (ii) answer queries about the decoder's distribution. As such, only the decoder needs to be tractable, and we can increase the expressivity of the continuation model by allowing the encoder to be an arbitrary neural network $q_{enc}$, giving rise to the following hybrid model:
\begin{equation}
    q_{\text{hybrid}}(x_{t+1:T}|x_{1:t}) = \sum_{z_t} q_{\text{enc}}(z_t | x_{1:t}) q(x_{t+1:T}|z_t)
    \label{eq:nnhmm-core}
\end{equation}
%As such, we rely on amortized inference: that is, predicting a (tractable) distribution $q^{(j)}$ for each context
%s such, we propose to take an amortized inference approach. In particular, we propose to use an probabilistic encoder-decoder architecture where the
%usina a probabilistic encoder $p_{\text{enc}}$ that maps the context $x_{1:t}^{(j)}$ to some latent state $z^{(j)}$, and the decoder is given by a conditional model $q(x_{t+1:T}|z^{(j)})$. In HMMs, the context $x_{1:t}$ and the continuation $x_{t+1:T}$ are conditionally independent given the latent $z_t$. Leveraging this property, we can amortize the conditional log-likelihood of the continuation by conditioning on a neural network generated $z_t$:
% A crucial difference of this compared to (for example) variational autoencoders is that we do not need to use variational inference as 
%If the encoder $q_{\text{enc}}(z_t | x_{1:t})$ is given by the HMM distribution $q(z_t|x_{1:t})$, then we recover the original HMM model of the continuation distibution. However, we observe this is unnecessarily restrictive: we only require the decoder $q(x_{1:t}|z_t)$ to be tractable (i.e., to be able to compute $q(\alpha|z_t)$), while we merely need to be able to evaluate $q_{\text{enc}}(z_t | x_{1:t})$.  
We can then jointly train the encoder and decoder to maximize the expected log-likelihood over a dataset of contexts and continuations $\{x_{1:t}^{(j)}, x_{t+1:T}^{(j)}\}_{j=1}^{N}$. By utilizing a more expressive encoder, our hypothesis is that the log-likelihood of the hybrid model will improve upon the pure HMM, in turn leading to better downstream performance.

We call our method Learning To Look Ahead (LTLA) as (i) we train a neural encoder to learn a language model's distribution over continuations, and (ii) we execute a symbolic algorithm to look ahead over exponentially many continuations with a probability distribution given by the output of the learned encoder. 
% Although we only maintain the model's tractability over the continuation tokens $x_{t+1:T}$, this is sufficient for downstream applications where tokens are generated autoregressively.
%to do tractable look-ahead over all possible continuations.
% \begin{equation}
%     \text{LL} = \sum_{j=1}^{N} \left[\log \sum_{z_t} q_{\text{enc}}(z_t | x_{1:t}^{(j)}) q(x_{t+1:n}^{(j)}|z_t)\right].
% \end{equation}
%Here the neural encoder $p_{\text{enc}}(z_t | x_{1:t})$ can be viewed as an expressive alternative to the HMM encoder $p_{\text{hmm}}(z_t | x_{1:t})$, when substituted for $p_{\text{enc}}$ in Equation~(\ref{eq:nnhmm-core}), recovers the original HMM formulation. Importantly, the 
Next, we describe the algorithms for and complexity of conditional probability queries of interest, and then discuss architectural choices for the neural~encoder. 

\begin{figure*}[t]
    \centering
    \includegraphics[width=\linewidth]{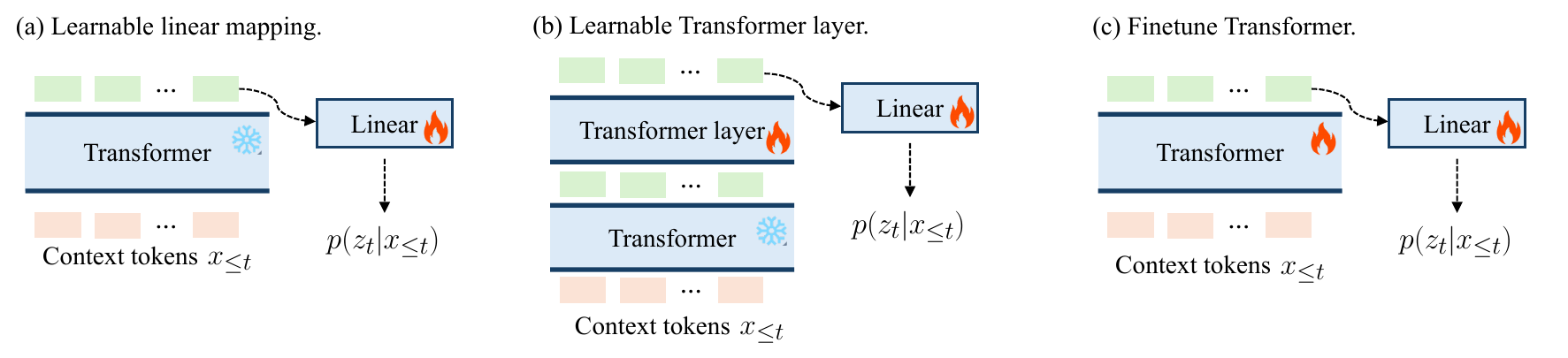}
    \vspace{-2em}
    \caption{Neural architectures for neural HMMs: (a) frozen Transformer with linear mapping, (b) frozen Transformer with additional learnable layer, (c) fully finetuned Transformer.}
    \label{fig:nhmm-structures}
    % \vspace{-0.8em}
\end{figure*}

\subsection{Inference using Neural HMMs}

After conditioning on a context $x_{\le t}$, we have a tractable model of the distribution of continuations $q(x_{>t}|x_{\le t})$, rendering many natural queries exactly and efficiently computable. 
% For example, after observing a sequence $x_{\le t}$, we might ask the following questions
% \begin{itemize}
%     \item What is the probability that the sequence ends with a ``Yes" or a ``True"?
%     \item What is the probability that the sequence ends within the next 10 tokens? 
%     \item What is the probability that the generated sequence will be labeled ``toxic" by a given a classifier?
% \end{itemize}
For example, we might ask 
(i) what is the probability that the sequence ends within the next $k$ tokens; 
(ii) what is the probability that the sequence contains a particular keyword;  
(iii) what is the probability that the generated sequence will be labeled ``toxic"/``biased"/etc by a given tractable classifier?
% \oliver{comment this paragraph from here on if space is needed....}
In general, queries $q(\alpha | x_{\le t})$, where $\alpha$ is some subset of the possible continuations $x_{>t}$, can be written \(q(\alpha | x_{\le t}) =\sum_{x_{>t}}q(x_{>t}|x_{\le t})\mathbbm{1}\{x_{>t}\in \alpha\}\) where $\mathbbm{1}\{x_{>t}\in \alpha\}$ is an indicator function of the membership of $x_{>t}$ in $\alpha$. So to compute $q(\alpha | x_{\le t})$, it suffices to \emph{multiply} $q(x_{>t} | x_{\le t})$ by the indicator function for $\alpha$ in such a way that the summation remains tractable; general sufficient conditions are known for such tractable products \citep{darwiche2002knowledge,zhang2025restructuring}. More generally, tractable models can support other types of queries, including for example information-theoretic quantities \citep{vergari2021compositional,wang2024compositional}.

\paragraph{Controlled Generation} As a concrete application of the tractability of a neural HMM, we consider the task of generating text conditioned on a (logical or semantic) constraint $\alpha$.
We sample autoregressively from the conditional distribution factored as in Equation \ref{eq:control}, where the term $p(x_t | x_{<t})$ is computed by the autoregressive model, and the term $p(\alpha | x_{\le t})$ is computed with the tractable neural HMM.
As an example, we describe the case where $\alpha$ is the event that the text is accepted by a given deterministic finite automaton (DFA);
% For logical constraints, we consider conditioning on regular expressions following \citet{ZhangNeurIPS24}, that is, the constraint that the text is accepted by a given deterministic finite automaton (DFA).
% For soft constraints, we consider attributes predicted by a factorized probabilistic classifier $p(s|x_{1:n})$, following \citet{YidouWengICML25}.
% To illustrate the use of neural-encoded HMMs for decoding in this manner, we focus on the case where $\alpha$ is the event that the text is accepted by a given DFA.
see \Cref{app:ctrlg-and-trace} for a full definition. 
We follow the Ctrl-G algorithm designed for unconditional HMMs \citep{ZhangNeurIPS24}.
% , adapting it slightly for a neural-encoded HMM.
% \paragraph{Ctrl-G} 
Given a DFA $M$, let $S_t$ be the random variable representing the state of $M$ after reading $x_{\le t}$ sampled from $p(x_{\le t}|\alpha)$. Then, 
\begin{align}\label{eq:ctrlg-neural-agg}
p(\alpha | x_{\le t})
% &=\sum_{z_t}p(z_t | x_{\le t})p(\alpha | z_t, x_{\le t})\\&
=\sum_{z_t}p(z_t | x_{\le t})p(\alpha | z_t, s_{t})
\end{align}
using the law of total probability together with the Markov properties of HMMs and DFAs (and the fact that $s_t$ is fully determined by $x_{\le t}$). 
The term $p(z_t | x_{\le t})$ may be estimated by the neural encoder. In practice, to avoid evaluating the neural encoder $V$ times, once for each candidate next token $x_t$, we instead evaluate the encoder once to compute $p(z_{t-1}|x_{<t})$ and then perform a single HMM forward step to obtain $p(z_{t}|x_{\le t})$ for each $x_t$.
Moreover, observe that the term $p(\alpha | z_t,s_t)$ is entirely independent of the context $x_{\le t}$, and so all $T \cdot V \cdot H$ such probabilities $p(\alpha | z_t,s_t)$ can be precomputed and stored in a lookup table before decoding via an efficient backward-style algorithm \citep{ZhangNeurIPS24}.
% Then to compute $p(\alpha| x_{\le t})$ for each candidate next token $x_t$, the $O(vh)$ forward probabilities $p(z_t|x_{\le t})$ are each computed via a single step of the standard HMM forward algorithm requiring time $O(\tau)$. 
% Finally $p(\alpha| x_{\le t})$ is computed via the aggregation in \Cref{eq:ctrlg-neural-agg}. 
% Thus $p(\alpha| x_{\le t})$ is, in total, computed by a single evaluation of the neural encoder, an $O(\tau)$-time HMM forward step, and an $O(vh)\in O(\tau)$-time aggregation in \Cref{eq:ctrlg-neural-agg}. 
% The full procedure is summarized in \Cref{app:ctrlg-and-trace}. 
% \Cref{alg:ctrlg-w-ltla}.
% \begin{wrapfigure}{R}{0.53\textwidth}
% \begin{minipage}{0.53\textwidth}
% \begin{algorithm}[H]%[H]
% \caption{Ctrl-G with LTLA}
% \label{alg:ctrlg-w-ltla}
% \begin{algorithmic}
% \Require $M = (Q, \Sigma, \delta, q_0, F)$, $p_{\LM}$, $p_{enc}$, $q$, $n$
% \For{$t \gets n$ down to $1$}
%     \State Pre-compute $q(\alpha \mid z_t, s_t)$ by \Cref{eq:ctrlg-backward-recurrence}
% \EndFor
% \State Initialize $s_0 \gets q_0$, $x_{1:0} \gets \emptyset$
% \For{$t \gets 1$ to $n$}
%     \State Compute $q_{\text{hybrid}}(\alpha \mid x_{\le t})$ by \Cref{eq:ctrlg-alpha-given-context}
%     \State Sample $x_t \propto q_{\text{hybrid}}(\alpha \mid x_{\le t}) \cdot p_{\LM}(x_t \mid x_{<t})$
%     \State Update $x_{\leq t} \gets x_{<t} \circ x_t$
%     \State Transition $M$: $s_t \gets \delta(s_{t-1}, x_t)$
% \EndFor
% \State \Return $x_{1:n}$
% \end{algorithmic}
% \end{algorithm}
% \end{minipage}
% \end{wrapfigure}
Similar approaches work for other constraints; TRACE was derived by \citet{YidouWengICML25} for the case where the constraint is an attribute $s$ predicted by a fully factorized probabilistic classifier $p(s|x_{1:n})$.
% .; here, the backward-style precomputation takes only time $O(n\tau)$.
% , i.e., having no asymptotic time increase over the standard HMM backward algorithm.

\paragraph{Complexity}
\label{sec:only latent conditional}
We emphasize an advantage of changing only the HMM \emph{encoder}.
Specifically, we compare with an alternative wherein a neural network predicts, given a context $x_{\le t}$, fresh parameters of a full HMM.
Then, the backward computation of $p(\alpha\mid z_t,s_t)$ can no longer be precomputed and must be carried out at each decoding step on the freshly predicted HMM, yielding decoding time that scales quadratically in sequence length.
Moreover, predicting fresh HMM parameters for each context induces a memory blow-up with batch size.
The resulting time and space complexities for training and inference are summarized in \Cref{app:complexities}.
A single HMM forward (or backward) step costs $O(H^2 + HV)$ for dense transition and emission~matrices.
%(always assuming $\tau =\Omega(HV)$)

% \guy{you need a table with design criteria/properties, and approaches and checkboxes for which ones have which property. Currently the requirements of what we want from this model is spread out between a footnote, a complexity paragraph, and scattered text elsewhere}

\subsection{Architecture}

We now specify our hybrid model architectural design: the HMM decoder and the neural network encoder.

% We now specify the architectural designs for our hybrid model, which includes two components: the HMM decoder and the neural network encoder.

\subsubsection{Structured Sparsity}
\label{sec:dense-vs-monarch}

The main parameter controlling the expressivity of an HMM is its \emph{hidden size}, the number of states of each latent variable. 
Specifically, for any index $t$ an HMM forms a Markov chain $X_{< t} \to Z_t\to X_{\ge t} $, and so any dependence of the continuation on the context must `flow through' the latent state.
Indeed, we can show that the \emph{logarithm} of the hidden size is an upper bound on the mutual information between the context and continuation.
Denoting the mutual information between $A$ and $B$ by $I(A;B)$, the entropy of $A$ by $H(A)$, and the support of $A$ by $\text{supp}(A)$, we have the following.
\begin{proposition}\label{prop:latent-bottleneck}
For any Markov chain  $X_{< t} \to Z_t\to X_{\ge t} $,
% $X_{ctx}\to Z\to X_{ctn}$, 
we have 
\(
I(X_{<t};X_{\ge t})\le H(Z_t)\le \log |\text{supp}\,(Z_t)|.
\)
\end{proposition}
This bound (proven in \Cref{app:proofs}) holds regardless of the encoder $p(z_t|x_{<t})$ or decoder $p(x_{\ge t}|z_t)$, i.e., it holds for standard HMMs as well as neural HMMs. Although Proposition \ref{prop:latent-bottleneck} provides a strong motivation for increasing the hidden size $H$ to improve the model's capacity, the number of parameters in an HMM with dense transition and emission matrices grows quadratically in hidden size. 
Therefore, we also consider using Monarch matrices \citep{dao2022monarch}, structured matrices that require polynomially fewer parameters in the hidden size yet can still express complex, high-rank matrices, which have been used for scaling tractable models \citep{ZhangICML25}.
Monarch matrices for the transition and emission in an HMM use $O(H^{3/2}+H^{1/2}V)$ parameters, yielding a single-step complexity of $O(H^{3/2}+H^{1/2}V)$.

\subsubsection{Neural Network Architectures}

Neural HMM performance depends on the encoder architecture, but downstream use also demands low overhead. We therefore reuse the base autoregressive backbone and add only lightweight modules to predict the HMM~prior.

Figure~\ref{fig:nhmm-structures} shows three variants. 
\textbf{(a) Linear head:} we freeze the Transformer backbone and learn a linear map from the last hidden state to the latent prior $p(z_t \mid x_{\le t})$. Since the LM hidden states are already computed during decoding, this adds negligible cost.
\textbf{(b) Extra layer:} we keep the backbone frozen and insert a small learnable Transformer layer before the linear head, increasing encoder capacity with modest overhead.
\textbf{(c) Finetune:} we finetune the transformer backbone jointly with the linear head, maximizing flexibility at the cost of training and inference overhead.

\begin{figure*}[t]
    \centering
    \includegraphics[width=0.9\linewidth]{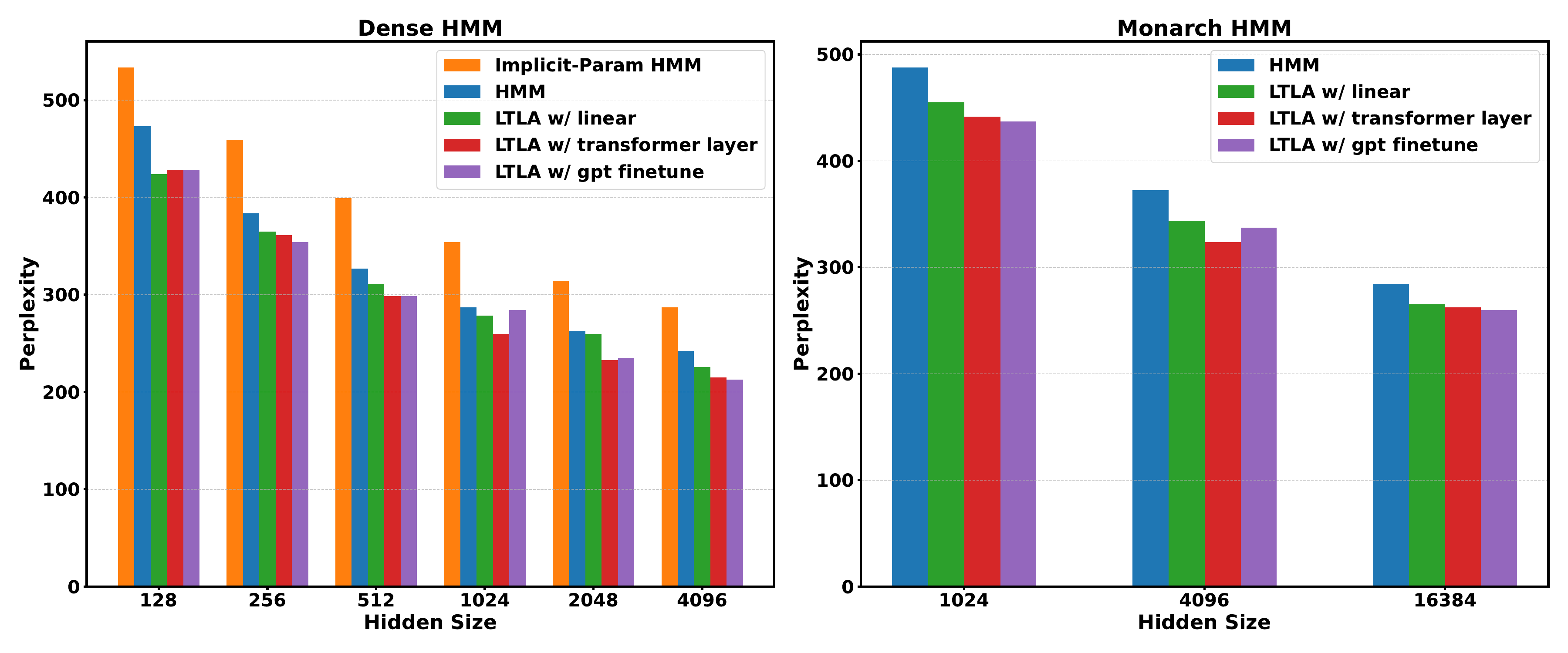}
    \vspace{-0.6em}
    \caption{Perplexity of neural HMMs and baseline HMM for varying hidden sizes with dense transition and emission matrices on the left, Monarch matrices on the right. 
    }
    % \guy{make more clear which ones are our contribution. maybe call them all LTLA(X)}\oliver{on the new plot, would be great to increase fontsize (messaged ian)}
    % }
    \label{fig:cll-results} 
\end{figure*}

\section{Experiments}

\vspace{-6pt}
We first assess \textsc{LTLA} as a tractable surrogate by how well it matches the base LM on continuation likelihoods. We then evaluate controlled generation under (i) \emph{syntactic} constraints (keyword/length DFAs) and (ii) \emph{semantic} constraints (detoxification via toxicity lookahead). \textsc{LTLA} distills a neural HMM from a fixed base generator (GPT-2 Large for text; Qwen2\mbox{-}VL\mbox{-}2B-Instruct for captioning).

\textbf{Baselines.}
For surrogate quality, we compare to a standard HMM and the approach of \citet{lee2025optimizing}. For controlled generation, we compare to prior decoding baselines for syntactic constraints and text detoxification, and to vanilla captioning and prompting for multimodal detoxification; a standard HMM controller isolates the effect of neural prefix encoding.

\textbf{Reproducibility.}
Complete training and decoding configurations, including task-specific data construction, hyperparameters, prompts, and qualitative examples, are provided in App.~\ref{app:exp-details} (GPT-2: Sec.~\ref{app:gpt2-exp}; Qwen2-VL: Sec.~\ref{app:qwen-exp}.

\textbf{Research questions.}
\textbf{(RQ1)} Does neural prefix encoding improve surrogate quality relative to a standard HMM?
\textbf{(RQ2)} How does encoder capacity affect this improvement?
\textbf{(RQ3)} Do better surrogates yield better controlled generation under syntactic and semantic constraints?
\textbf{(RQ4)} Can \textsc{LTLA} condition effectively on multimodal context?

\begin{table}[t]
\small
\centering
\caption{\textbf{CommonGen.} HMM-based methods guarantee 100\% constraint satisfaction, and \textsc{LTLA} improves generation quality (BLEU-4, ROUGE-L, and CIDEr) over a standard HMM.}
\label{tab:ctrlg-constraint}
\setlength{\tabcolsep}{4pt}
\begin{tabular}{lcccc}
\toprule
\textbf{Method} & \textbf{BLEU-4} & \textbf{ROUGE-L} & \textbf{CIDEr} & \textbf{Constraint} \\
\midrule
FUDGE        & 0.246 & 0.404 & --   & 47.0\% \\
A*esque      & 0.282 & 0.434 & 1.52 & 98.8\% \\
\textbf{Standard HMM} & 0.290 & 0.438 & 1.55 & \textbf{100.0\%} \\
\textbf{LTLA} & \textbf{0.321} & \textbf{0.454} & \textbf{1.68} & \textbf{100.0\%} \\
\bottomrule
\end{tabular}
\end{table}

\subsection{Distillation Performance \& Analysis}

Figure~\ref{fig:cll-results} reports the scaling behavior of neural HMMs compared to standard HMM baselines. The a-axis refers to the hidden size of the HMM. Across both dense and Monarch parameterizations (as introduced in Section~\ref{sec:dense-vs-monarch}), augmenting HMMs with neural components consistently reduces perplexity relative to the baseline. Across the three neural architectures, adding a trainable Transformer layer (i.e., Figure~\ref{fig:nhmm-structures}(b)) or finetuning GPT (i.e., Figure~\ref{fig:nhmm-structures}(c)) achieves stronger performance compared to the learnable linear layer. These results demonstrate that neural HMMs scale favorably with hidden dimension, leading to better conditional log-likelihood per token/perplexity. We further observed that the neural encoder achieves better perplexity particularly for shorter continuation lengths, which is to be expected as the first few tokens have the strongest dependence on the context and are where the neural encoder has the most direct effect; see \Cref{fig:stratified_cll} in \Cref{app:stratified} for a plot showing perplexity for varying continuation lengths.
% In Figure \ref{fig:stratified_cll}, we further see that the neural encoder achieves better perplexity particularly for shorter continuation lengths, which is to be expected as the first few tokens have the strongest dependence on the context and are where the neural encoder has the most direct effect.

In terms of the architectural choices, both adding a transformer layer and finetuning the full GPT model lead to greater improvements in perplexity compared to using a linear head, at the cost of additional compute. 
In terms of the HMM architecture, the results show that, perhaps surpisingly, the larger hidden sizes of Monarch HMMs did not perform significantly better when normalized for compute. For instance, the Monarch HMM with size $16384$ uses theoretical compute comparable to the Dense HMM with hidden size $1024$, but does not show significantly better perplexity.

%MI between token being generated, and continuation. 

For another reference point, we evaluated the optimization baseline of \citet{lee2025optimizing}, which combines the implicit parameterization of \citet{chiu2020scaling} with latent-variable distillation (LVD) \citep{LiuICLR23}; in Fig.~\ref{fig:cll-results} (left) this corresponds to the \textsc{Implicit-Param HMM} curve/bar. In our distillation setting, it does not outperform a carefully tuned standard HMM, likely reflecting the sensitivity of standard HMM training to AdamW hyperparameters, and it remains well behind our neural HMMs. %Secondly, a line of work has considered enhancing the expressivity via squaring. We normalize for compute. 
% basically, they were using a bad initialization strategy for HMM parameters and a bad weight decay value that caused a plateau. fixing these meant that it outperformed neural reparameterization. 
% also, beta = 0.95 instead of 0.999

% requires: \usepackage{caption}  % for \captionsetup

\begin{table*}[t]
\centering
\small

\begin{minipage}[t]{0.68\textwidth}
\captionsetup{width=\linewidth}
\caption{{Detoxification for text-only GPT-2 (RealToxicityPrompts) and Qwen2-VL (Hateful Memes).}\footnotemark\ $^\dagger$ marks methods with unusually low PPL that also exhibit low diversity, consistent with mode-collapsed ``safe'' continuations; see \S\ref{paragraph:semantic-constraints}. Full baseline tables are in App.~\ref{app:detox}.}
\label{tab:detox-summary}
\vspace{-0.5em}
\begin{tabular}{llcccc}
\toprule
\textbf{Setting} & \textbf{Method} & \textbf{Max tox} ($\downarrow$) & \textbf{Avg tox} ($\downarrow$) & \textbf{dist-2} ($\uparrow$) & \textbf{PPL} ($\downarrow$) \\
\midrule
\multirow{7}{*}{GPT-2 text}
  & GPT2 baseline   & 0.385 & 0.254 & \textbf{0.87} & \textbf{25.57} \\
  & GeDi            & 0.363 & 0.217 & 0.84 & 60.03 \\
  & FUDGE           & 0.302 & 0.371 & 0.78 & 12.97$^\dagger$ \\
  & PPLM            & 0.520 & 0.518 & 0.86 & 32.58 \\
  & DPO             & 0.180 & 0.026 & 0.76 & 21.59$^\dagger$ \\
  & \textbf{Standard HMM} & 0.163 & 0.016 & 0.85 & 29.83 \\
  & \textbf{LTLA}   & \textbf{0.152} & \textbf{0.015} & 0.85 & 29.81 \\
\midrule
\multirow{3}{*}{Qwen2-VL}
  & Vanilla         & 0.087 & 0.249 & 0.45 & \textbf{2.36} \\
  & Prompting       & 0.078 & 0.224 & 0.46 & 2.46 \\
  & \textbf{LTLA}   & \textbf{0.064} & \textbf{0.188} & \textbf{0.48} & 3.56 \\
\bottomrule
\end{tabular}
\end{minipage}
\qquad
\begin{minipage}[t]{0.23\textwidth}
\captionsetup{width=\linewidth}
\caption{Inference-time overhead on GPT-2 detoxification.}
\label{tab:detox-overhead}
\vspace{0.6em}
\setlength{\tabcolsep}{4pt}
\begin{tabular}{lc}
\toprule
\textbf{Method} & \textbf{Ratio} \\
\midrule
Base LM & 1.00 \\
Prompting & $\sim$3.0 \\
GeDi / DExperts & 2.0--3.0 \\
Mix and Match & 7.5 \\
MuCoLa & 15--20 \\
PPLM & 40.0 \\
\textbf{LTLA} & \textbf{1.14} \\
\bottomrule
\end{tabular}
\end{minipage}

\vspace{-0.8em}
\end{table*}

\subsection{Controlled Generation}

Our neural HMMs both predict future behavior more accurately, which improves benchmark performance, and plug in easily to other models such as VLMs where many existing control methods (including standalone HMMs) are difficult to use. We demonstrate versatility across two constraint families, \emph{hard logical constraints} and \emph{soft semantic attributes}, using established constrained-decoding frameworks \citep{ZhangNeurIPS24,YidouWengICML25} for both LLMs and VLMs (details in App.~\ref{app:exp-details}). By learning to look ahead more accurately, we will show that LTLA improves constraint satisfaction, fluency and quality.

\textbf{Logical constraints.}
We enforce hard constraints (e.g., keyword constraints) by combining the tractable lookahead model with deterministic finite automata (DFAs) \citep{ZhangNeurIPS24}. We distill both standard HMMs and \textsc{LTLA} variants from a GPT-2 model finetuned on CommonGen, construct DFAs that accept only sequences satisfying the constraints, then run constrained beam search and select the top hypothesis by base-LM log-likelihood. We evaluate using BLEU-4, CIDEr, ROUGE-L, and perplexity. As shown in Table~\ref{tab:ctrlg-constraint}, HMM-based methods are the only ones that guarantee constraint satisfaction (100\% versus 47.0--98.8\% for prior controllable baselines such as FUDGE and A*esque). Beyond satisfaction, \textsc{LTLA} improves BLEU/ROUGE/CIDEr over a standard HMM and substantially reduces average and maximum perplexity (See App~\ref{tab:ctrlg_len20}, \ref{app:ctrlg_ppl32}), suggesting that the neural prior provides more context-aware guidance under hard constraints without forcing unnatural sequences.

\vspace{-10pt}
\paragraph{Semantic constraints in LLMs and VLMs.}
\label{paragraph:semantic-constraints}
For semantic control, we couple the tractable lookahead model with a log-linear toxicity classifier trained on RealToxicityPrompts and scored with the Perspective API, following \citet{YidouWengICML25}. We distill \textsc{LTLA} from the base model (GPT-2 Large for text; Qwen2\mbox{-}VL\mbox{-}Instruct\mbox{-}2B for images). During decoding, the surrogate’s predicted future toxicity provides a lookahead cost that reweights the next-token distribution.

On \emph{text-only detoxification}, we evaluate GPT-2 Large on RealToxicityPrompts and compare against representative controllable-generation baselines, including trained guides (GeDi, FUDGE), logit-control (PPLM), RL-style approaches (DPO), and the HMM-based controller. On \emph{multimodal detoxification}, we apply the same mechanism to Qwen2-VL captioning on Hateful Memes, comparing against vanilla captioning and prompt engineering.

Table~\ref{tab:detox-summary} summarizes the main results (full baseline tables are in App.~\ref{app:detox}). On GPT-2, \textsc{LTLA} reduces maximum toxicity from 0.254 to 0.015 while keeping fluency and diversity comparable to the base model. Some RL-style or guide-based methods report very low perplexity, sometimes even lower than the base model, but this often coincides with sharply reduced diversity, suggesting ``overly fluent'' mode-collapsed continuations rather than genuinely higher-quality generations \citep{holtzman_curious_2020}. Relative to the standard HMM controller, \textsc{LTLA} further improves toxicity while preserving a similar fluency/diversity profile.

On Qwen2-VL, \textsc{LTLA} reduces maximum toxicity from 0.249 to 0.188 and improves diversity, albeit with higher perplexity. We additionally report long-sequence detoxification in App.~\ref{app:detox} (1k-token generations with 128-step lookahead), where \textsc{LTLA} continues to reduce toxicity while keeping fluency metrics nearly stable, indicating that the benefits extend to longer contexts.

\textbf{Minimal Overhead.}
\textsc{LTLA} reuses the LM hidden state (no extra LM calls) and uses a fixed-horizon HMM with cached backward messages, so decoding adds only a vectorized one-step HMM update per token. Empirically on GPT-2 detoxification, this yields a near-constant per-token overhead: \textsc{LTLA} is $1.14\times$ the base LM, while guidance-style baselines are substantially slower (often $2$--$40\times$; Table~\ref{tab:detox-overhead}).

% \begin{figure}
%     \centering
%     \includegraphics[width=0.5\linewidth]{}
%     \caption{Figure showing overhead of each method over the base generation}
%     \label{fig:time}
% \end{figure}
% \subsection{VLM}

% As a demonstration of the capabilities....

\section{Related Work and Conclusion}

Future-dependent conditional queries generalize next-token prediction and make the LM continuation distribution the key object of interest \citep{boyd2022predictive}. Since exact conditioning is typically intractable, prior work approximates it via amortized inference, energy-based decoding-time inference, or sequential Monte Carlo over continuations \citep{hu2023amortizing,qin2022cold,lew2023sequential,zhao2024probabilistic,loula2025syntactic,ahmed2025semantic}. In parallel, tractable probabilistic models aim to make such conditional inference exact by construction and are actively being scaled in practice \citep{ProbCirc20,LiuICML24,chiu2020scaling,lee2025optimizing}. Our contribution is architectural: we learn a \emph{globally trained} HMM lookahead surrogate and condition only its latent prior on frozen transformer embeddings, aligning with hybrid neural--tractable modeling. \citep{shao2020conditional,dos2024probabilistic}.

% Tractable probabilistic models in general constitute a growing research area with interesting theoretical \citep{} and practical challenges \citep{}.
% and propose a method based on tractable probabilistic inference \citep{ProbCirc20}
% \citep{qin2022cold,hu2023amortizing,lew2023sequential,krause2020gedi,yang2021fudge,meng2022controllable}

% HMMs:
% \begin{itemize}
%     \item \citep{chiu2020scaling} Chiu \& Rush. Neural reparam. Blocked emissions (similar to monarch). 
% \end{itemize}
% Other generative models
% normalizing flows

% \section{Conclusions}

% In conclusion, w
We proposed Learning to Look Ahead (LTLA), a method for \emph{learning tractable continuation distributions} that support efficient future-dependent queries while remaining compatible with modern transformer representations. By separating lookback (LM embeddings) from lookahead (a tractable decoder) and learning a prefix-conditioned latent prior, LTLA improves surrogate quality and transfers these gains to downstream control under syntactic and semantic constraints. More broadly, LTLA strengthens the case that continuation modeling can be both \emph{accurate and tractable}, turning the LM lookahead distribution into an explicit object we can learn, query, and use for probabilistic reasoning.

\clearpage
\section{Impact Statement}
This paper presents work whose goal is to advance the field of machine learning. There are many potential societal consequences of our work, none of which we feel must be specifically highlighted here.

\bibliography{icml2026_conference}
\bibliographystyle{icml2026}

\newpage 
\newpage 

\appendix

\section{Proofs}\label{app:proofs}

We provide a proof of \Cref{prop:latent-bottleneck}, restated here for convenience.

\begin{proposition}
For any Markov chain  $X_{< t} \to Z_t\to X_{\ge t} $,
% $X_{ctx}\to Z\to X_{ctn}$, 
we have 
\[
I(X_{<t};X_{\ge t})\le H(Z_t)\le \log |\text{supp}\,(Z_t)|.
\]
\end{proposition}

\emph{Proof.}
We have $I(X_{< t};X_{\ge t})\le I(X_{<t};Z_t)=H(Z_t)-H(Z_t\mid X_{<t} )\le H(Z_t)$.
The first inequality is the `data processing inequality' (e.g., proved via the chain rule of mutual information), the equality is a standard identity that follows from definitions, and the final inequality holds because entropies are nonnegative.\hfill $\square$
% \qedsymbol

% \section{Hidden Markov Models}\label{app:hmms}

% A Hidden Markov Model (HMM) represents a joint distribution over variables $X_1,X_2,\ldots,X_n$ with latent variables $Z_1,Z_2,\ldots,Z_n$. Specifically, the distribution of an HMM is given by 
% \[
% p(x_1,\ldots,x_n)=\sum_{z_1,\ldots,z_n}p(z_1)p(x_1|z_1)\prod_{t=2}^np(z_t|z_{t-1})p(x_t|z_{t-1}).
% \]
% \Cref{fig:hmm} shows an HMM as a graphical model.

% \begin{figure}[h]\centering
% \begin{tikzpicture}
% \tikzstyle{node}=[draw,circle,minimum size=22,inner sep=0]
% \coordinate (dx) at (1.4,0);
% \coordinate (dy) at (0,-1.3);

% \node(z1) [node] at (0,0) {$Z_1$};
% \node(z2) [node] at ($(z1)+(dx)$) {$Z_2$};
% \node(z3) [node] at ($(z2)+(dx)$) {$Z_3$};
% \node(z4) [] at ($(z3)+(dx)$) {$\ldots$};
% \node(zn) [node] at ($(z4)+(dx)$) {$Z_n$};

% \node(x1) [node] at ($(z1)+(dy)$) {$X_1$};
% \node(x2) [node] at ($(x1)+(dx)$) {$X_2$};
% \node(x3) [node] at ($(x2)+(dx)$) {$X_3$};
% \node(x4) [] at ($(x3)+(dx)$) {$\ldots$};
% \node(xn) [node] at ($(x4)+(dx)$) {$X_n$};

% \draw [->] (z1) -- (z2);
% \draw [->] (z2) -- (z3);
% \draw [->] (z3) -- (z4);
% \draw [->] (z4) -- (zn);
% \draw [->] (z1) -- (x1);
% \draw [->] (z2) -- (x2);
% \draw [->] (z3) -- (x3);
% \draw [->] (zn) -- (xn);

% \end{tikzpicture}
% \caption{Hidden Markov Model as a graphical model.}
% \label{fig:hmm}
% \end{figure}

\section{DFAs}\label{app:ctrlg-and-trace}

We give a formal definition of a deterministic finite automaton.

% To supplement our description of Ctrl-G, here we provide a formal definition of a DFA as well as psuedocode summarizing the procedure described in the main text.

\begin{definition}
A deterministic finite automaton (DFA) is a tuple $M= (Q, \Sigma, \delta, q_0, F)$, where $Q$ is a
finite set of states, $\Sigma$ a finite set of symbols, $\delta:Q\times \Sigma \to Q$ a transition function,
$q_0$ an initial state, and $F\subseteq Q$ a set of accept states. A string of tokens $w_1w_2\ldots w_n$ is accepted by
$M$ if there exists a sequence of states $q_0,q_1,\ldots,q_n$ such that $q_{i}=\delta(q_{i-1}
, w_{i})$ for $1\le i\le n$ and $q_n \in F$.
\end{definition}

\section{Complexities}\label{app:complexities}
Time and space complexities of training and constrained generation algorithms with neural HMM variants are given in \Cref{tab:complexities}.
Here, $n$ is the decoded sequence length, $B$ is batch size, $H$ is the number of HMM states, $V$ is vocabulary size, and $m$ is beam size. We write $\tau$ for the time of a single HMM forward (or backward) step and $s$ for the cost of one neural-encoder forward pass that predicts the latent prior from the prefix representation.

\bgroup
\def\arraystretch{1.2}%  1 is the default, change whatever you need
\begin{table*}[h]
\centering
\begin{tabular}{r r | l l l}
         &       & Standard            & Neural        & Full Conditioning   \\
         &       & HMM                 & HMM                   & HMM                 \\
\hline
Training & Time  & $O(Bn\tau)$         & $O(Bn(\tau +s))$      & $O(Bn(n\tau +s))$   \\ 
         & Space & $O(\tau + nBH)$     & $O(\tau + nBH)$       & $O(Bn\tau )$        \\ 
\hline
Ctrl-G   & Time  & $O(n\tau m)$      & $O(n(\tau m+s))$      & $O(n(n\tau m+s))$  \\ 
         & Space & $O(\tau + nmH)$     & $O(\tau + nmH)$       & $O(nVm)$           \\  
TRACE    & Time  & $O(n\tau)$          & $O(n(\tau +s))$       & $O(n(n\tau +s))$   \\
         & Space & $O(\tau + nH)$      & $O(\tau + nH)$        & $O(n\tau )$        \\

% with batch size
% Ctrl-G   & Time  & $O(Bn(\tau m+s))$   & $O(Bn(\tau m+s))$     & $O(Bn(n\tau m+s))$  \\ 
%          & Space & $O(n(Vm+B))$        & $O(n(Vm+B))$          & $O(n(Vm+B))$        \\  
% TRACE    & Time  & $O(Bn(\tau +s))$    & $O(Bn(\tau +s))$      & $O(Bn(n\tau +s))$   \\
%          & Space & $O(n(V+B))$         & $O(n(V+B))$           & $O(Bn\tau )$        \\
\end{tabular}
\caption{
Time and space complexities of training and constrained generation algorithms with neural HMM variants.
The parameters are generation length $n$; vocabulary size $V$; batch size $B$; HMM hidden size $H$; HMM single-step complexity $\tau$; number of edges $m$ in the DFA (for Ctrl-G); and time $s$ for a single evaluation of the neural model.
}
\label{tab:complexities}
\end{table*}
\egroup

\section{Stratification by Continuation Length}\label{app:stratified}

\begin{figure}[h!]
    \centering
    \centering
    \includegraphics[width=0.85\linewidth]{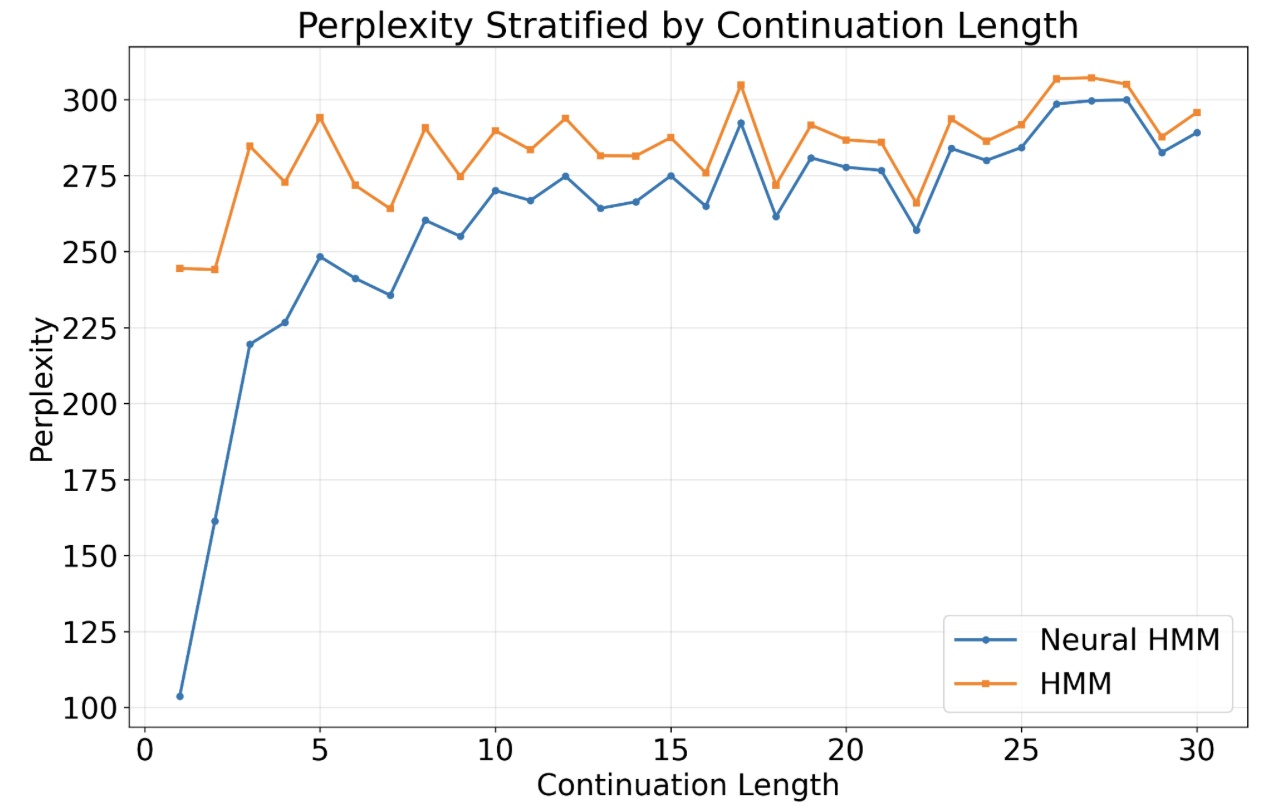}
    \caption{Perplexity of (Monarch) HMMs vs neural HMMs on the GPT2-large dataset for different continuation lengths. } % \guy{are the lines in the legend swapped? neural HMM has worse perplexity than HMM}
    \label{fig:stratified_cll}
    % \begin{subfigure}{0.46\linewidth}
    % \includegraphics[width=0.95\linewidth]{figs/mutual_information.png}
    % \caption{
    % The mutual information $I(X_{\le t};X_{t+k})$ between the context $X_{\le t}$ and the $k$th token of the continuation $X_{t+k}$ for varying $k$. 
    % % In all methods, mutual information is much higher for earlier tokens in the continuation. Neural-encoding and Monarch matrices (separately or together) produce an increase in mutual information.
    % }
    % \label{fig:placeholder}
    % \end{subfigure}
\end{figure}

See \Cref{fig:stratified_cll}.

\section{Experimental Details}
\label{app:exp-details}

This appendix provides further details on the LTLA training setups and decoding configurations omitted from the main text. We describe the GPT-2 language-model experiments (Sec.~\ref{app:gpt2-exp}) and the Qwen2-VL vision--language experiments (Sec.~\ref{app:qwen-exp})).

\subsection{GPT-2 Neural HMM Training and Evaluation}
\label{app:gpt2-exp}

\paragraph{Training data construction.}
We train three task-specific GPT-2 \textsc{LTLA} surrogates, each distilled from a different base generator and data source:
\begin{itemize}
    \item \textbf{Surrogate-quality model (likelihood assessment).} We distill from GPT-2 Large using \emph{unconditional} generations of length 32. We form context--continuation pairs by choosing a split index uniformly at random and treating the prefix as context and the suffix as continuation.
    \item \textbf{Syntactic-control model (CommonGen).} We distill from GPT-2 Large finetuned on CommonGen, using \emph{unconditional} generations of length 32 from the finetuned model. We construct context--continuation pairs using the same uniform-random split.
    \item \textbf{Semantic-control model (RealToxicityPrompts).} We distill from GPT-2 Large using \emph{conditional} generations from RealToxicityPrompts training prompts. For each prompt, we sample a continuation of up to 32 tokens and choose a split index uniformly at random within the generated sequence to form the context--continuation pair.
\end{itemize}
Unless otherwise noted, all GPT-2 results use the corresponding task-specific surrogate above.

\paragraph{Data sizes.}
Unless otherwise noted, we use:
\begin{itemize}
    \item \textbf{Training set:} 10M generated sequences;
    \item \textbf{Test set:} 10k generated sequences (disjoint from training).
\end{itemize}

\paragraph{Objective and optimization.}
All GPT-2 HMM variants (standard and \textsc{LTLA}) are trained to maximize the conditional log-likelihood of the continuation under the HMM,
\(
\log p_\theta(x_{>T} \mid z_T)
\),
using standard negative log-likelihood loss with teacher forcing. Unless specified otherwise, we use:
\begin{itemize}
    \item Optimizer: AdamW,
    \item Learning rate: $2\times 10^{-2}$,
    \item Batch size: 256,
    \item Gradient accumulation: 1,
    \item Number of epochs: chosen so that each model sees at least $10^8$ continuation tokens.
\end{itemize}

\paragraph{LTLA encoder architectures.}
All \textsc{LTLA} models use GPT-2 Large as a frozen context encoder. We consider two families of neural heads.

\subsubsection*{GPT-2 Linear-Only Head (\texttt{gpt2\_linearonly})}
This is the default \textsc{LTLA} configuration used in our main experiments. GPT-2 weights are frozen; only a single projection layer is trained to map the last hidden state to a prior over HMM states.

Given a context sequence $x_{\le T}$, we extract the last hidden state
\[
\mathbf{h}_T
=
\mathrm{GPT2}(x_{\le T})_{T}
\in \mathbb{R}^{1280},
\]
and compute the \textsc{LTLA} prior
\begin{align*}
\log p(z_T \mid x_{\le T})
=
\mathrm{logsoftmax}\bigl(W \mathbf{h}_T + \mathbf{b}\bigr),
\\
W \in \mathbb{R}^{H \times 1280},
\end{align*}
where $H$ is the number of HMM hidden states (typically $H=4096$ for dense HMMs, or the smallest perfect square $\ge V$ when using Monarch-structured transitions and emissions).
Only $W$ and $\mathbf{b}$ are trained; GPT-2 parameters remain frozen.

\subsubsection*{GPT-2 Block Head with Cross-Attention (\texttt{gpt2\_block})}
For a higher-capacity encoder, we also experiment with adding a small cross-attention block on top of GPT-2's frozen hidden states.

Given the full sequence of hidden states
\(
H = (\mathbf{h}_1,\dots,\mathbf{h}_T)
\),
we treat the last position as a query and the entire prefix as keys/values:
\[
\tilde{\mathbf{h}}_T = \mathrm{CrossAttnBlock}(\mathbf{h}_T, H),
\]
where the block consists of two stacked layers of the form
\begin{align*}
x \leftarrow x + \mathrm{CrossAttn}(\mathrm{LN}(x), \mathrm{LN}(H)), \\
x \leftarrow x + \mathrm{MLP}(\mathrm{LN}(x)).
\end{align*}
We then map $\tilde{\mathbf{h}}_T$ to the \textsc{LTLA} prior via a linear layer and log-softmax, as in \texttt{gpt2\_linearonly}. Again, GPT-2 itself is frozen; only the cross-attention block and final projection are trained.

\paragraph{HMM parameterization.}
For all GPT-2 experiments we use either dense or Monarch-structured transitions and emissions:
\begin{itemize}
    \item \textbf{Dense HMM:} transition and emission matrices in $\mathbb{R}^{H \times H}$ and $\mathbb{R}^{H \times V}$ with row-wise softmax to enforce stochasticity.
    \item \textbf{Monarch HMM:} structured low-parameter matrices where the number of states $H$ is chosen as the smallest perfect square $\ge V$ to match the Monarch factorization.
\end{itemize}
The HMM parameters are updated jointly with the neural head using the optimizer and learning rate above.

\subsection{Qwen2-VL Neural HMM Setup}
\label{app:qwen-exp}

\paragraph{Base vision--language model.}
For the multimodal detoxification experiments we use Qwen2-VL-2B-Instruct (\texttt{Qwen/Qwen2-VL-2B-Instruct}) as the base model. We treat Qwen2-VL as a black-box encoder--decoder; LTLA only requires access to its hidden states and token logits.

\paragraph{Caption sampling for training data.}
To construct training pairs for the neural HMM, we generate captions for images from the Hateful Memes training split using Qwen2-VL with the prompt
\[
\text{\small User: [IMAGE] Describe this image.}
\]
and the following decoding parameters:
\begin{itemize}
    \item Max new tokens: 32,
    \item Temperature: 0.8,
    \item Top-p: 0.9,
    \item Generations per image: 1.
\end{itemize}
We then choose a split index uniformly at random within the generated caption and treat the prefix as context and the suffix as continuation. To fit GPU memory, we store only the last hidden state at the split position as input to the neural HMM encoder.

\paragraph{Neural HMM.}
For Qwen2-VL we use a Monarch-structured HMM with:
\begin{itemize}
    \item Hidden size: $H = 152{,}100$ (smallest perfect square $\ge$ vocabulary size plus special tokens),
    \item Neural head: single linear layer from the Qwen2-VL hidden dimension (1536) to $\mathbb{R}^{H}$, followed by log-softmax to obtain $\log p(z_T \mid x_{\le T})$.
\end{itemize}

\paragraph{Training hyperparameters.}
The Qwen2-VL HMM is trained with negative log-likelihood on the continuation tokens, using:
\begin{itemize}
    \item Optimizer: AdamW,
    \item HMM learning rate: $1\times 10^{-2}$,
    \item Neural head learning rate: $1\times 10^{-5}$,
    \item Batch size: 180.
\end{itemize}

\paragraph{Decoding configuration for detoxification.}
At evaluation time on Hateful Memes, we prompt Qwen2-VL with:

\medskip
\small User: [IMAGE] Describe this image. Avoid toxic, offensive, or hateful language while preserving descriptive accuracy.

\normalsize
and decode using:
\begin{itemize}
    \item Sampling: top-p nucleus sampling with $p=0.9$,
    \item Temperature: $1.0$,
    \item Max new tokens: 32,
    \item Generations per image: 25.
\end{itemize}
The HMM-based controller (standard or \textsc{LTLA}) is queried at each step with a fixed lookahead horizon as described in the main text.

% \subsection{Qualitative Detoxification Examples}
% \label{app:qual-examples}

% We include a representative example from Hateful Memes to illustrate the effect of \textsc{LTLA} on multimodal detoxification. We do not reproduce the original meme images to avoid amplifying hateful content; instead we provide textual descriptions.

% \paragraph{Example.}
% \begin{itemize}
%     \item \textbf{Image:} Person wearing a headscarf and hoop earrings, smiling at the camera; the meme text contains a slur.
%     \item \textbf{Baseline Qwen2-VL caption:}
%     \emph{``The image shows a person wearing a headscarf and hoop earrings, smiling. The caption reads: `a head diaper is required when you have shit for brains'.''}
%     \item \textbf{LTLA-controlled caption:}
%     \emph{``The image depicts a meme featuring a smiling individual wearing a headscarf, with the text `a head scarf is required when you have no skills' above them.''}
% \end{itemize}

% The baseline model reproduces offensive text from the meme, whereas \textsc{LTLA} produces a paraphrased, non-toxic description while preserving the overall structure and intent of the caption.

\clearpage
\section{Additional Results on CommonGen}
\label{app:ctrlg}
See \Cref{tab:commongen-results,tab:ctrlg_len20,app:ctrlg_ppl32}.

\begin{table}[h]
\centering
\caption{CommonGen results on GPT2-large. Higher BLEU/ROUGE/CIDEr/SPICE is better ($\uparrow$); higher constraint satisfaction is better ($\uparrow$).}
\label{tab:commongen-results}
\begin{tabular}{lcccccccccc}
\toprule
 & \multicolumn{2}{c}{BLEU-4 ($\uparrow$)} & \multicolumn{2}{c}{ROUGE-L ($\uparrow$)} &
   \multicolumn{2}{c}{CIDEr ($\uparrow$)}  & \multicolumn{2}{c}{SPICE ($\uparrow$)} &
   \multicolumn{2}{c}{Constraint ($\uparrow$)} \\
\cmidrule(lr){2-3} \cmidrule(lr){4-5} \cmidrule(lr){6-7} \cmidrule(lr){8-9} \cmidrule(lr){10-11}
 & \textit{dev} & \textit{test} & \textit{dev} & \textit{test} &
   \textit{dev} & \textit{test} & \textit{dev} & \textit{test} &
   \textit{dev} & \textit{test} \\
\midrule
\rowcolor[gray]{0.93}
\multicolumn{11}{l}{\textit{supervised} \;-- base models trained with full supervision} \\
FUDGE   & -     & 0.246 & -     & 0.404 & -    & -    & -     & -     & -        & 47.0\% \\
A*esque & -     & 0.282 & -     & 0.434 & -    & 1.52 & -     & 0.308 & -        & 98.8\% \\
NADO    & 0.308 & -     & 0.444 & -     & 1.61 & -    & 0.320 & -     & 88.8\%   & -      \\
\midrule
\rowcolor[gray]{0.93}
\multicolumn{11}{l}{\textit{unsupervised} \;-- base models not trained with keywords as supervision} \\
standard HMM       & 0.303 & 0.290 & 0.443 & 0.438 & 1.56 & 1.55 & 0.302 & 0.303 & \textbf{100.0\%} & \textbf{100.0\%} \\
\textbf{LTLA} & \textbf{0.320} & \textbf{0.321} & \textbf{0.453} & \textbf{0.454} &
\textbf{1.63} & \textbf{1.68} & - & - & \textbf{100.0\%} & \textbf{100.0\%} \\
\bottomrule
\end{tabular}
\end{table}

\begin{table}[th]
\centering
\caption{Additional Ctrl-G results on CommonGen with max sequence length 20. Higher BLEU/CIDEr/ROUGE-L is better ($\uparrow$); lower perplexity is better ($\downarrow$).}
\label{tab:ctrlg_len20}
\scalebox{0.9}{
\begin{tabular}{lccccc}
\toprule
\textbf{Model} & \textbf{BLEU-4} ($\uparrow$) & \textbf{CIDEr} ($\uparrow$) & \textbf{ROUGE-L} ($\uparrow$) & \textbf{Avg.\ PPL} ($\downarrow$) & \textbf{Max PPL} ($\downarrow$)\\
\midrule
Standard HMM & 0.301 & 1.552 & \textbf{0.448} & 39.59 & 1569.51 \\
HMM with Linear NN & \textbf{0.303} & \textbf{1.566} & \textbf{0.448} & 41.99 & 1458.38\\
HMM with Transformer Block & 0.297 & 1.536 & 0.446 & \textbf{36.24} & \textbf{616.73}\\
\bottomrule
\end{tabular}}
\end{table}

\begin{table}[th]
\centering
\small
\caption{Additional Ctrl-G results on CommonGen with max sequence length 32. Higher BLEU/CIDEr/ROUGE-L is better ($\uparrow$); lower perplexity is better ($\downarrow$).}
\label{app:ctrlg_ppl32}
\vspace{-0.8em}
\scalebox{0.9}{
\begin{tabular}{lccccc}
\toprule
\textbf{Model} & \textbf{BLEU-4} ($\uparrow$) & \textbf{CIDEr} ($\uparrow$) & \textbf{ROUGE-L} ($\uparrow$) & \textbf{Avg.\ PPL} ($\downarrow$) & \textbf{Max PPL} ($\downarrow$)\\
\midrule
Standard HMM & 0.303 & 1.566 & 0.448 & 36.00 & 1569.51\\
LTLA: HMM with Linear NN & 0.311 & 1.566 & 0.448 & 34.16 & \textbf{671.88}\\
LTLA: HMM with Transformer Block & \textbf{0.320} & \textbf{1.625} & \textbf{0.453} & \textbf{33.98} & 1065.47 \\
\bottomrule
\end{tabular}}
\end{table}

\clearpage
\newpage
\section{Additional Detoxification Results}\label{app:detox}
See \Cref{tab:rtp_full,tab:rtp_long}.

\begin{table}[th]
\centering
\small
\caption{Full GPT-2 Large detoxification results on RealToxicityPrompts. Lower toxicity and PPL are better; higher dist-2/3 is better.}
\label{tab:rtp_full}
\resizebox{\textwidth}{!}{%
\begin{tabular}{lcccccl}
\toprule
\textbf{Model} &
\multicolumn{2}{c}{\textbf{Toxicity ($\downarrow$)}} &
\multicolumn{2}{c}{\textbf{Diversity ($\uparrow$)}} &
\textbf{PPL ($\downarrow$)} &
\textbf{Approach Type} \\
& \textbf{avg.\ max} & \textbf{prob.} & \textbf{dist-2} & \textbf{dist-3} & & \\
\midrule
\rowcolor[gray]{0.93} \multicolumn{7}{l}{\textbf{GPT-2 Large Results}} \\
GPT2 & 0.385 & 0.254 & 0.87 & \textbf{0.86} & \textbf{25.57} & Baseline \\
DAPT\textsuperscript{(1)} & 0.428 & 0.360 & 0.84 & 0.84 & 31.21 & Finetuning \\
GeDi\textsuperscript{(2)} & 0.363 & 0.217 & 0.84 & 0.83 & 60.03 & Decoding (Trained Guide) \\
FUDGE\textsuperscript{(3)} & 0.302 & 0.371 & 0.78 & 0.82 & 12.97 & Decoding (Trained Guide) \\
DExperts\textsuperscript{(4)} & 0.314 & 0.128 & 0.84 & 0.84 & 32.41 & Decoding (Trained Guide) \\
PPLM\textsuperscript{(5)} & 0.520 & 0.518 & 0.86 & 0.86 & 32.58 & Decoding (Logit Control) \\
MuCoLa\textsuperscript{(6)} & 0.308 & 0.088 & 0.82 & 0.83 & 29.92 & Decoding (Sampling) \\
PPO\textsuperscript{(7)} & 0.218 & 0.044 & 0.80 & 0.84 & 14.27 & RL \\
Quark\textsuperscript{(8)} & 0.196 & 0.035 & 0.80 & 0.84 & 12.47 & RL \\
DPO\textsuperscript{(9)} & 0.180 & 0.026 & 0.76 & 0.78 & 21.59 & RL \\
TRACE (HMM) & 0.163 & 0.016 & 0.85 & 0.85 & 29.83 & Decoding (HMM Reasoning) \\
\textbf{LTLA} & \textbf{0.152} & \textbf{0.015} & 0.85 & 0.85 & 29.81 & Decoding (Neural HMM Reasoning) \\
\bottomrule
\end{tabular}%
}
\end{table}

\begin{table}[th]
\centering
\caption{Long-sequence detoxification on RealToxicityPrompts with GPT-2 Large (1k-token continuations, 128-step LTLA lookahead). Lower toxicity and PPL are better; higher dist-2/3 is better.}
\label{tab:rtp_long}
\resizebox{0.8\textwidth}{!}{%
\begin{tabular}{lccccc}
\toprule
\textbf{Method} & \textbf{Max tox} ($\downarrow$) & \textbf{Avg tox} ($\downarrow$) & \textbf{PPL} ($\downarrow$) & \textbf{dist-2} ($\uparrow$) & \textbf{dist-3} ($\uparrow$) \\
\midrule
GPT-2 baseline      & 0.369 & 0.226 & 13.74 & \textbf{0.787} & \textbf{0.948} \\
LTLA (Neural HMM)   & \textbf{0.156} & \textbf{0.006} & \textbf{13.65} & 0.774 & 0.943 \\
\bottomrule
\end{tabular}%
}
\end{table}

\end{document}

%% file: math_commands.tex
\usepackage{amsmath,amsfonts,bm}

\def\eqref#1{equation~\ref{#1}}
\def\1{\bm{1}}

\DeclareMathAlphabet{\mathsfit}{\encodingdefault}{\sfdefault}{m}{sl}
\SetMathAlphabet{\mathsfit}{bold}{\encodingdefault}{\sfdefault}{bx}{n}

%% file: macros.tex
\newcommand{\ctx}{x_{\text{ctx}}}

\newcommand{\ctn}{x_{\text{ctn}}}